\documentclass[letterpaper]{article} 
\usepackage{aaai21}  
\usepackage{times}  
\usepackage{helvet} 
\usepackage{courier}  
\usepackage[hyphens]{url}  
\usepackage{graphicx} 
\usepackage{natbib}  
\usepackage{caption} 
\usepackage[english]{babel}
\usepackage[utf8x]{inputenc}
\usepackage{amsmath}
\usepackage{parskip}
\usepackage{tikz}
\usepackage{amssymb}
\usepackage{amsthm}
\usepackage{booktabs}
\usepackage{amsfonts}
\usepackage{relsize}
\usepackage{enumitem}
\usepackage{afterpage}
\usepackage{multirow}
\usepackage{hyperref}

\usepackage{breakurl}

\title{An Experimental Study of the Transferability of Spectral Graph Networks}								
\author{Axel Nilsson}								
\date{9 November 2020}											

\makeatletter
\let\thetitle\@title
\let\theauthor\@author
\let\thedate\@date
\makeatother

\author{%
  Axel Nilsson and Xavier Bresson\\
}
\affiliations{
School of Computer Science and Engineering\\
Nanyang Technological University (NTU), Singapore \\
contact@xelnilsson.com, xbresson@ntu.edu.sg
}

\begin{document}

\maketitle

%

\begin{abstract}
Spectral graph convolutional networks are generalizations of standard convolutional networks for graph-structured data using the Laplacian operator.
A common misconception is the unstability of spectral filters, i.e. the impossibility to transfer spectral filters between graphs of variable size and topology. This misbelief has limited the development of spectral networks for multi-graph tasks in favor of spatial graph networks.
However, recent works have proved the stability of spectral filters under graph perturbation.
Our work complements and emphasizes further the high quality of spectral transferability by benchmarking spectral graph networks on tasks involving graphs of different size and connectivity. Numerical experiments exhibit favorable performance on graph regression, graph classification and node classification problems on two graph benchmarks. The implementation of our experiments are available on GitHub for reproducibility.  

\end{abstract}

\section{Introduction}
Graph neural networks \cite{scarselli_graph_2009} is the class of networks that process data on graphs. Convolutional neural networks \cite{lecun1998gradient}, which have shown great performances on a variety of tasks defined on Euclidean domains such as computer vision,
have been extended to graphs with spectral theory \cite{bruna_spectral_2014, defferrard_convolutional_nodate} and spatial template matching \cite{kipf_semi-supervised_2017}. Graph convolutional networks (GCNs) have showed significant performances in myriads of domains, among others the representation of social networks to describe communities \cite{kipf_semi-supervised_2017}, fake news detection \cite{monti_fake_2019},
chemistry \cite{gilmer_neural_2017}, 
knowledge graphs \cite{schlichtkrull_modeling_2017, hamilton_inductive_2017},
physics \cite{cranmer_learning_2019}, and
recommendation systems \cite{monti_geometric_nodate, ying_graph_2018}.


In this paper, we focus on the transferability of spectral GCNs. Spectral GCNs define learnable filters as parametric functions of the graph Laplacian operator.
Transferability is an essential property of learning systems, related to their generalization capability. It demonstrates the ability of the system to learn consistent and discriminative features that can be used to make prediction for graphs unseen by the model during training.
\cite{levie_transferability_2019,levie_transferability_2019-1} disproved the idea that spectral filters cannot be transfered to different graphs. Precisely, the authors in \cite{levie_transferability_2019} showed that given a graph filter $g$ and a small perturbation $E$ with $\|E \|\leq 1$ on the graph Laplacian $\Delta$ then the filter on the perturbed graph is a small perturbation of the original filter:
$$
\| g(\Delta) - g(\Delta+E) \| = O(\|E \|).
$$

In other words, the perturbation of the filter is only bounded by the perturbation of the graph, and filters are thus stable. Combining stability with equivariance, graph spectral filters are proven to be transferable.
Besides, in \cite{levie_transferability_2019-1}, they showed the robustness of spectral filters w.r.t. small graph perturbations by assuming the graphs are discretized from the same “continuous” space. They established the transferability error of a given filter $g$:
$$
\textrm{Err}(g) \leq \textrm{Err}(\Delta) \ + \ \textrm{Err}(\textrm{Consistency}).
$$

The transferability error is thus bounded by the perturbation of the Laplacian $\Delta$ and the consistency error that vanishes for large graphs. Spectral GCNs are thus robust if graphs are discretized from the same underlying space.
Because of the misconception of failure of spectral filter transfer, the use of spectral GCNs in a multi-graph setting has been scarce in the literature. \cite{knyazev_spectral_2018} applied spectral GCNs to a set of multiple molecular graphs and \cite{ktena2018metric} to brain connectivity networks.

The goal of this work is to validate the theoretical results on the transferability of spectral filters with a corpus of experimental evidence. Specifically, we will show that ChebNets \cite{defferrard_convolutional_nodate} performs favorably for the fundamental tasks of graph classification, graph regression and node classification using two graph benchmarks. Another  class of spectral networks is CayleyNets \cite{levie2018cayleynets}, which may also be investigated in a future work.





\section{ChebNets \cite{defferrard_convolutional_nodate}}
These graph networks define smooth spectral filters $g_{\theta}$ parametrized with Chebyshev polynomials $T_i$ applied to the normalised Laplacian operator $\Delta$:
\begin{equation}
   \Delta =I - D^{-\frac{1}{2}} A D^{-\frac{1}{2}},
\end{equation}
where $A$ is the adjacency matrix and $D$ is the degree matrix. The spectral filters $g_{\theta}$ are defined as
\begin{equation}
    g_{\theta} (\tilde{\Delta})h= \sum_{i=0}^k \theta_i T_i(\tilde{\Delta}) h,
		\label{eq:learnedfilterscheb}
\end{equation}
where $k$ is the number of Chebyshev polynomials, $\theta$ are the learnable parameters, $h$ is a signal defined on the graph, and $\tilde{\Delta} = 2\lambda_{\textrm{max}}^{-1}\Delta − I_n$ is the Laplacian re-normalized such that its eigenvalues are in the interval $[-1,1]$, where the Chebyshev polynomials are orthogonal. A great computational advantage of these parametric filters is to define them with a recursive equation:
\begin{equation}
    \begin{cases} T_0 = h \\T_1 = \tilde{\Delta}T_0 \\ T_{k\geq2} = 2\tilde{\Delta}T_{k-1}-T_{k-2} \end{cases}
    \label{eq:recursT}
\end{equation}
which reduces the computational complexity to $O(n)$ for sparse graphs.

\section{Graph Benchmarks}

Recent projects \cite{hu_open_2020, dwivedi_benchmarking_2020-1} have proposed benchmarks with collections of datasets to evaluate and compare GNNs. They not only provide openly accessible and peer-reviewed datasets but also leaderboards, helping to track the performance of different GNN models.
In this work, we will use datasets from the Open Graph Benchmark (\textsc{ogb} version 1.2.3) \cite{hu_open_2020} and Benchmarking-GNNs \cite{dwivedi_benchmarking_2020}.
The selected datasets are composed of multiple graphs and cover a variety of tasks such as graph regression (\textsc{Zinc} , ogbg-Molpcba), graph classification (ogbg-Molhiv) and node classification (\textsc{cluster}, \textsc{Pattern}). A summary of these datasets and some of their proprieties can be found in Table \ref{tab:experimentsSummary}.
We notice that three of the five selected datasets are made of molecular graphs. This is simply because there exist only a few real-world datasets  of multiple graphs with varying sizes. Nonetheless, they exhibit significant differences, in respect of task and size, making them complementary.

\begin{table*}[htb]
\centering
\begin{tabular}{@{}lllll@{}}
\toprule
         & \# graphs &  \# nodes  & Problem type    & Metric \\ \midrule
\textsc{Cluster}  &  12'000   & 41-190     & Node classification &  \multirow{2}{*}{Weighted Acc}      \\
\textsc{Pattern}  &  14'000   & 44-188     & Node classification &      \\
\textsc{Zinc}     & 12'000    & 9-35       & Graph regression    &  \textsc{mae}      \\
ogbg-Molhiv           &   41'127  & 25.5*      & Graph regression    & \textsc{roc-auc} \\
ogbg-Molpcba          & 437'929   & 26.0*      & Graph regression    & \textsc{AP}\\ \bottomrule
\end{tabular}
\caption{Summary of the different tasks with the number of graphs, problem type and type of metric. *is the average node number instead of the range.}
\label{tab:experimentsSummary}
\end{table*}

\subsection{\textsc{Pattern \& Cluster}}

\textsc{Pattern} and \textsc{Cluster} are node classification datasets for graphs generated synthetically with stochastic block models (SBMs) \cite{abbe_community_2017}, commonly used to model communities in social networks.
Each node has an intra-probability of being connected to a node in the same communitie and an extra-probability of being connected to a node in other communities.
For \textsc{Cluster}, the task is to identify communities in a semi-supervised setting. Each graph has six SBM communities of 5 to 35 nodes, and one node in each community is labeled at random.
For \textsc{Pattern}, the task is pattern matching, that is recognizing pre-defined sub-graphs embedded in larger graphs. There are 100 randomly generated patterns of 20 nodes with random features.
The train/test/val splitting is 10K/2K/2K respectively for \textsc{Pattern} and 10K/1K/1K for \textsc{Cluster}.
More details on the construction of these datasets can be found in the paper \cite{dwivedi_benchmarking_2020}.

\subsection{\textsc{Zinc}}
\textsc{Zinc} is a graph regression dataset composed of molecular graphs. The task is to predict the constrained solubility of each molecule \cite{jin2018junction}, a continuous variable. The accuracy is determined by the mean average error of the $L1$ loss over the test set.
Node and edge features are categorical and correspond to the type of atoms and bonds.
The number of graphs in the training/validation/test sets are respectively 10K/1K/1K graphs.

\subsection{\textsc{Ogbg-Mol}}

ogbg-Molhiv and ogbg-Molpcba are two molecular datasets from the \textsc{ogb} benchmark, used for regression tasks.
ogbg-Molhiv is the smallest of the two datasets with 41,127 graphs.
The task is to predict if a given molecule would inhibit the replication of the HIV, cast as a binary label. It is evaluated by a \textsc{roc-auc} performance metric.
ogbg-Molpcba represents a more difficult task as the set is larger with 437,929 graphs, and there are 128 properties to regress for each graph. Besides, the class distribution is skewed with only 1.4\% of positive data. The performance is evaluated with average precision (AP). Both datasets are split by scaffolding to split the graphs based on their structure. This contrasts with \textsc{Zinc} which is randomly split, making easier to generalize \cite{hu_open_2020}.

\section{ChebNet Architectures}

Although it is essential to compare the performance of GCNs to contextualize a new model, there is no perfect approach to compare two GCNs.
Therefore, each benchmark has defined a set of rules to enable the comparison of models. In this work, we will follow the strategy of each benchmark to compare ChebNets with the following popular models; GCN \cite{kipf_semi-supervised_2017}, GraphSage \cite{hamilton_inductive_2017}, GAT \cite{velickovic_graph_2018} and GIN \cite{xu_how_2019}.

All experiments are run on a single GeForce RTX 2070 8GB GPU. The source code is available at:
\url{https://github.com/Axeln78/Transferability-of-spectral-gnns}.

\subsection{Benchmarking-GNNs}
In Benchmarking-GNNs \cite{dwivedi_benchmarking_2020}, a budget of 100,000 learnable parameters with four convolutional layers are fixed to compare the performance with the same learning capacity. We will follow this constraint when designing ChebNets for \textsc{Cluster}, \textsc{Pattern} and \textsc{Zinc}. 
The learning hyperparameters are the same as in \cite{dwivedi_benchmarking_2020} to prevent improving the performance by tweaking. That is, the batch size is 128 graphs, Adam optimizer has an initial learning rate of $10^{-3}$, reduced by a factor of 0.5 if the validation accuracy does not decrease every 5 epochs, batch normalisation and no dropout. Finally, the score is averaged over four runs with pre-selected seeds.

\subsection{\textsc{Ogbg-Mol}}
For the selected \textsc{ogb} datasets, ogbg-Molhiv and ogbg-Molpcba, we will adhere to the default architectures described in \cite{hu_open_2020} i.e. those used for GIN and GCN.
They are composed of an Atom- and Edge- Embedding layer, five convolutional layers, a mean-pooling layer with a hidden dimensionality of 300 followed by three linear layers with a tuned dropout ratio $\in \{0.0 , 0.5\}$.
The learning hyperparameters are the default \textsc{ogb} values, with a learning rate of $10^{-3}$ for the Adam optimizer and a batch size of 128. The result is the average over ten runs with random seeds.
A summary of the ChebNet architectures for all datasets is available in Table. \ref{tab:models}.
Finally, observe that the \textsc{ogb} leaderboard has no strict rules about model architecture, parameter budget, and hyperparameter selection.
Therefore we will only provide the two models GIN and GCN, which have similar architectures as a basis of comparison, and discuss more informally models that are on the leaderboard as of this writing.

We also notice that both benchmarks do not have any spectral GCNs to compare to in their leaderboard.

\subsection{Values of $k$ and $\lambda_{\textrm{max}}$}
We select the number of Chebyshev polynomials to be $k=5$ for \textsc{Pattern} and \textsc{Cluster}. This value is related to the size of the communities (5-35 nodes). For \textsc{Zinc}, the chosen Chebyshev order is $k = 2$ as the graphs are small (9 to 35 nodes). 
For ogbg-Molhiv and ogbg-Molpcba, the order of polynomial is selected to be $k=3$.

For all ChebNets, the spectral parameter $ \lambda_{\textrm{max}} = 2$ is fixed, as the largest eigenvalues can sometimes be numerical unstable to compute on certain graphs. Besides, fixing $ \lambda_{\textrm{max}} = 2$ is mathematically justified as the spectrum of the normalized Laplacian is bounded by this value \cite{chung1997spectral}. Another advantage is to reduce the computational cost and allow to compare with papers using this approximation \cite{knyazev_image_2019, knyazev_spectral_2018}.

\begin{table*}[ht!]
\centering
\begin{tabular}{@{}llc@{}}
\toprule
Dataset           &   Model Architecture                     & Hyperparam.          \\ \midrule
\textsc{Cluster}  & 7 -E70 -ChN70 -ChN70 -ChN70 -ChN70 -MP70 -L35 -L17 -L6 & $k=5$ \\
\textsc{Pattern}  & 3 -E70 -ChN70 -ChN70 -ChN70 -ChN70 -MP70 -L35 -L17 -L2 & $k=5$ \\
\textsc{Zinc}     & 28 -E106 -ChN106 -ChN106 -ChN106 -ChN106 -MP106 -L53 -L26 -L1 (No-RC) & $k=2$\\
ogbg-Molhiv  &   -AE300 -ChN300 -ChN300 -ChN300 -ChN300 -ChN300 - MP300 -L150 -L75 -L1 & $k=3$   \\
ogbg-Molpcba &   -AE300 -ChN300 -ChN300 -ChN300 -ChN300 -ChN300 - MP300 -L150 -L75 -L128 & $k=3$ \\ \bottomrule
\end{tabular}
\caption{Summary of all model architectures used. E stands for Embedding, AE for AtomEmbedding,  MP for Mean Pooling, L for Linear, ChN for ChebNets layer, and No-RC for no residual connections used. Each layer type is followed by a number indicating the output dimension.}
\label{tab:models}
\end{table*}

\section{Numerical Results}

Overall, numerical experiments show that ChebNets perform favorably well compared to other popular GCN models with comparable architectures. Table \ref{table:summary} reports the performance of ChebNets for each task.

\begin{table*}[ht!]
\centering
\centering
\begin{tabular}{@{}llrcl@{}}
\toprule
Dataset                      & Model     & \# parameters & Accuracy                                      & Metric                                                \\ \midrule
\multirow{5}{*}{\textsc{ Cluster} }    & \textbf{ChebNet}   &  102,535       & \textbf{73.13 $\pm$ 0.64  }    & \multirow{5}{*}{Weighted accuracy in \% (higher is better)} \\
                             & GAT*       & 110,700        & $ 57.73 \pm 0.32  $                            &                                                       \\
                             & GIN*       & 100,884       & $ 49.64 \pm 2.09  $                             &                                                       \\
                             & GCN*       & 103,077       & $47.82 \pm 4.91 $                              &                                                       \\
                             & GraphSage* & 99,139        & $ 44.89 \pm 3.70  $                             &                                                       \\ \midrule
\multirow{5}{*}{\textsc{ Pattern} } & \textbf{ChebNet}   & 102,183        & \textbf{85.75 $\pm$ 0.02 }    &  \multirow{5}{*}{Weighted accuracy in \% (higher is better)}  \\
                              & GIN*       & 100,884       & $85.59 \pm 0.01 $           &   \\
                             & GraphSage* & 98,607        & $78.20 \pm 3.06 $                              &                                                       \\
                             & GAT*       & 109,936        & $ 75.82 \pm 1.82  $                            &                                                       \\
                             & GCN*       & 100,923       & $74.36 \pm 1.59  $                             &                                                       \\ \midrule
\multirow{5}{*}{\textsc{ Zinc} } & \textbf{ChebNet}   & 101,230       & \textbf{0.360  $\pm$ 0.028}   &         \multirow{5}{*}{MSE (lower is better)}                \\
                              &  GIN*       & 103,079       & $ 0.387 \pm 0.015$       &                                                   \\
                             & GCN*       & 103,077       & $0.459 \pm 0.006$                             &                                                       \\
                             & GraphSage* & 94,977        & $0.468 \pm 0.003$                             &                                                       \\
                             & GAT*       & 102,385        & $0.475 \pm 0.007$                            &                                                       \\ \midrule
\multirow{3}{*}{ogbg-Molhiv} & \textbf{ChebNet}   & 1,465,351    & \textbf{0.7631 $\pm$ 0.0127} & \multirow{3}{*}{ROC-AUC (higher is better)}           \\
                             & GCN*       & 527,701       & $0.7606 \pm 0.0097$                           &                                                       \\
                             & GIN*       & 1,885,206     & $ 0.7558 \pm 0.0140$                          &                                                       \\ \midrule
\multirow{3}{*}{ogbg-Molpcba} & \textbf{ChebNet} & 1,475,003 & \textbf{0.2306 $\pm$ 0.0016} & \multirow{3}{*}{Average precision (AP) (higher is better)} \\
                             & GIN*       & 1,923,433     & $0.2266 \pm 0.0028$                           &                                                       \\
                             & GCN*       & 565,928       & $0.2020 \pm 0.0024$                           &                                                       \\
                             \bottomrule
\end{tabular}
\caption{Numerical study of ChebNets on several datasets. The models are ranked w.r.t. performance. The proposed ChebNets are in bold. The performances of the models identified with * are taken from the leaderboards of different benchmarks as of Nov. 9th, 2020. The results are the average and standard deviation over four runs for tasks from Benchmarking-GNNs and ten for OGB. ChebNets show favorable performance compared to other models by achieving the best performance in all five tasks.}
\label{table:summary}
\end{table*}

\subsection{\textsc{Cluster}}
ChebNets outperform all models for the \textsc{Cluster} dataset and perform at the level of deeper models found in the leaderboard of the benchmark.
In fact, the strong performance implies that ChebNets hold a good inductive bias for identifying community clusters in a semi-supervised node classification setting. This was expected as Laplacian-based clustering techniques have showed significant performance for community detection \cite{von2007tutorial}. This is in contrast to the current literature as the a commonly used model for this task has been GCNs \cite{kipf_semi-supervised_2017}, which perform  worse (25 percentage points).
Furthermore, if we consider that this dataset is a set of discretizations of a continuous manifold, then this result is a salient confirmation of the theory of \cite{levie_transferability_2019-1}.

\subsection{\textsc{Pattern}}
ChebNets perform better than the other models for \textsc{Pattern} and performs closely to GIN. Both models are have significantly higher accuracy than the rest. Although is expected that GIN performs well on tasks of pattern and graph matching, it is striking that the ChebNet performs equally well. Additionally, we observe no significant improvement in accuracy with deeper models in the benchmark leaderboard for this task.

\subsection{\textsc{Zinc}}
The \textsc{Zinc} experiment allows to compare ChebNets with a range of models with a similar budget of learnable parameters. Comparing to the leaderboard in \cite{dwivedi_benchmarking_2020}, ChebNets not only performs significantly better than spatial GCN models, they exhibit superior performances over models with the same depth although some are using edge features like GAT or Gated GCN \cite{bresson_residual_2018}.

\subsection{\textsc{ogbg-mol}}


For the two \textsc{ogb} tasks, ChebNets achieve better performance than the two models of comparison, GIN and GCN.
Unexpectedly, ChebNets outperform GIN, a model designed with maximal representation power w.r.t. the Weisfeiler-Lehman graph isomorphism test \cite{weisfeiler1968reduction}, while using 25\% less parameters.
Finally, comparing our results with the \textsc{ogb} leaderboard also reveals that ChebNets provide the best performance over the other reported GCNs that do not consider additional data augmentation techniques.

Overall the results show that ChebNets provide an efficient architecture for molecule tasks with \textsc{Zinc}, ogbg-Molhiv and ogbg-Molpcba. The presented results can be further improved by modifying the architecture to consider the edge bond features, which are important information about molecular structure.

To summarize, it is clear that the performance of ChebNets  is consistently high compared to the most used GCN architectures. This delivers a significant experimental proof that ChebNets work well on datasets of multiple graphs with variable sizes and for different tasks.

\section{Conclusion}
This experimental work investigates the transferability capacity of spectral GCNs on two graph benchmarks.
Numerical experiments demonstrate that ChebNets perform better than most popular spatial GCNs with comparable parameter budgets. These numerical experiments strongly support recent analytical results \cite{levie_transferability_2019,levie_transferability_2019-1} that spectral GCNs can compete at least as well as other spatial GCNs in the multi-graph setting. Such results are promising to encourage the development of new spectral networks.

\section{Acknowledgments}
XB is supported by NRF Fellowship NRFF2017-10.

\bibliography{references}
\end{document}